\documentclass[runningheads]{llncs}
\usepackage{graphicx}
\newcommand{\ra}[1]{\renewcommand{\arraystretch}{#1}}
\usepackage[ruled,vlined,linesnumbered]{algorithm2e}
\begin{document}
\title{A Lexicon and Depth-wise Separable Convolution Based Handwritten Text Recognition System}

\author{Lalita Kumari\inst{1}, Sukhdeep Singh\inst{2}, VVS Rathore\inst{3} \and
Anuj Sharma\inst{1}}

\institute{Deparment of Computer Science and Applications, Panjab University, India
\email{\{lalita,anujs\}@pu.ac.in}\\
\url{https://anuj-sharma.in}\and
D.M. College(Aff. to Panjab University, Chandigarh),Moga, India\\
\email{sukha13@ymail.com}\and
Physical Research Laboratory, Ahmedabad, India\\
\email{vaibhav@prl.res.in}}

\maketitle              
\begin{abstract}
Cursive handwritten text recognition is a challenging research problem in the domain of pattern recognition. The current state-of-the-art approaches include models based on convolutional recurrent neural networks and multi-dimensional long short-term memory recurrent neural networks techniques. These methods are highly computationally extensive as well model is complex at design level. In recent studies, combination of convolutional neural network and gated convolutional neural networks based models demonstrated less number of parameters in comparison to convolutional recurrent neural networks based models. In the direction to reduced the total number of parameters to be trained, in this work, we have used depthwise convolution in place of standard convolutions with a combination of gated-convolutional neural network and bidirectional gated recurrent unit to reduce the total number of parameters to be trained. Additionally, we have also included a lexicon based word beam search decoder at testing step. It also helps in improving the the overall accuracy of the model. We have obtained 3.84\% character error rate and 9.40\% word error rate on IAM dataset; 4.88\% character error rate and 14.56\% word error rate in George Washington dataset, respectively. 

\end{abstract}
\section{Introduction}
Handwritten Text Recognition (HTR) is a complex and widely studied computer vision problem in research community. In HTR, cursive strokes of handwritten text need to be recognized. Available text can be either in online or offline form \cite{Kumari2022}. In online HTR, the time ordered sequence of pen tip is captured. While in offline HTR, static images of handwritten text is available. In this work, we have focused on offline HTR. Handwriting recognition, specially offline HTR systems pose challenges, such as, variability of strokes not only varying among writers but also in a single writer as well, poor and degraded quality of available document images, slop and slant present in text, variable inter (in between lines) and intra (in between characters) spaces among the letters that need to be recognised and limited availability of labelled dataset needed for training of HTR model. Modern deep learning based techniques are used to solve this complex task. \par Initially, Hidden Markov Models (HMM) based techniques are used to solve HTR problems. In this technique, a text image is pre-processed using various computer vision techniques and hand crafted features as aspect ratios of individual characters are manually extracted from an image and fed to HMM based classifiers for recognition. Due to HMMs limited capacity of extracting contextual information and manual feature selection, recognition results are poor. In last few decades, deep learning based methods are primarily used for this task. Convolutional Neural Networks (CNN), Recurrent Neural Networks (RNN), Convolutional Recurrent Neural Networks (CRNN), Gated-CNN, Multi-Dimensional Long Short-Term Memory Recurrent Neural Networks (MDLSTM-RNNs) are some state-of-the-art machine learning techniques used to solve HTR problem. Convolution based techniques used to extract fine features of input image and RNN based techniques provides memory to remember long character sequences. Connectionist Temporal Classification (CTC) is widely used to train and test these Neural Network (NN) based systems in an end-to-end manner. Since, the emergence of artificial intelligence and machine learning based devices in day-to-day life,
latest research trends in HTR domain are favoured for a robust, less complex system with less number of trainable parameters and acceptable accuracy. In this work, we have proposed a HTR model that is favoured in this direction as well. We have applied depth-wise separable convolutional operation in place of standard convolution to reduce number of trainable parameters. A combination of convolutions, depth-wise convolutions and Full-Gated convolutions along with BGRU units are used to recognised text lines of benchmarked datasets in an end-to-end manner. Following are the key contributions of our present study,
\begin{itemize}
	\item A novel end-to-end HTR system is given to recognize text lines.
	\item Proposed model is able to improve recognition accuracy with less number of trainable parameters (Number of training parameters 820,778).
	\item Recognition text is lexically confined with the help of Word Beam Search (WBS) \cite{Scheidl2018} decoder.
	\item Overall pipeline of HTR system including essential steps are presented for better understanding of HTR system.
	\item We have achieved 3.84\% Character Error Rate (CER) and 9.40\% Word Error Rate (WER) on IAM dataset and 4.88\% CER and 14.56\% WER in George Washington (GW) dataset, respectively. These state-of-the-art results achieved using lesser training parameters. 
\end{itemize}
Extensive experiments are performed on benchmarked datasets such as IAM. The rest of the paper is organised as follows, section 2 include the key contributions in text line recognition. Section 3 demonstrates proposed architecture. Extensive experiments are presented in section 4. Results obtained reported in section 5 and section 6 include the conclusion of the present study.

\section{Related Work}
In this section, previous works in the domain of HTR has been discussed. We have focused upon text line HTR recognition task. This section present the genesis of HTR that helps readers to understand evaluation of significant HTR techniques. At start, researchers tackle HTR task by using Dynamic Programming (DP) based approaches on word level images, where optimum path  finding based algorithm is used. In DP based approaches although, character accuracy improves but this does not guarantee the improvement in the overall word accuracy \cite{chen99,bellman2015}. Later, HMM based techniques used in HTR \cite{2002vin}. Combination of N-Gram language models and optical recognition using HMM with Gaussian mixture emission probability (HMM-GMM) is one of the primarily studied HTR technique \cite{Toselli2004,SANCHEZ2019}. Optical models of HMMs are further improved by using Multi Layer perceptron (MLP) as emission probability \cite{zamora2011,dreuw2011} and discriminative training techniques in HMM-GMMs \cite{Toselli2015}. The MLP is constrained by fixed length input, thus LSTM-HMM based techniques are used in HTR system \cite{doetsch2014,kozielski2013}. The HMMs have drawbacks such as, probability of each observation depends only on current state, therefore, limited contextual information is less utilised. Apart from that HMMs are generative in nature while discriminative performed better in classification and labelling tasks \cite{Liwicki2012}. \par As a alternative to HMMs, Recurrent Neural Networks (RNNs) does not suffer from these limitations. A model which is having only RNN, it works on character level only because objective function of NN require separate training signal at each point of input signal \cite{nikolaos95}. The Connectionist Temporal Classification (CTC) is a RNN output layer used in sequence labelling task \cite{graves2006}. Models trained with the help of CTC do not require pre segmented data and able to provide probability distribution of label sequence. The variation of LSTM such as BLSTM and Multidimensional LSTM (MDLSTM) are used with CTC to give state-of-the art recognition results\cite{Liwicki2012,cirricum2013,pham2014,bluche2016,Voigtlaender2016}. The LSTM and BLSTM work in one dimensional sequences and MDLSTM captures long term dependencies across the both directions, but it is highly computationally expensive. Later, it was shown that similar results can be acquired by stacking few more BLSTM layers \cite{puigcerver2017}. One more popular architecture, that is, Convolutional Recurrent Neural Network (CRNN) is first introduced in scene-text recognition task \cite{shi2017}. It consist of stack of convolutional layers followed by one or more layers of BLSTM and softmax output layers, which include occurrence probability of N (numbers of characters in dataset) +1 (CTC blank) tokens \cite{SimpleHtr2018}. A compact and fast model compare to \cite{puigcerver2017} was proposed using Gated CNN architecture \cite{bluche2017}. A combination of both the model is proposed by applying Gated-CNN in \cite{puigcerver2017} model and able to get promising results on benchmarked datasets \cite{flor2020}. In one such study, attention based techniques clubbed with CNN and BLSTM  are also used by researchers to recognize cursive text \cite{Doetsch2016}. Similarly, a attention based end-to-end system with internal segmentation for paragraph recognition system is studied \cite{Coquenet2022}. Table \ref{table:tab1} summarizes all the related work and their techniques.
\begin{table}[!htbp]
	\ra{1.2}
	\caption{Summary of techniques used in Handwritten Text Recognition}
	
	\hskip-3em	\begin{tabular}{ll}
		
		\hline
		\textbf{S.NO.} & \textbf{Techniques} \\ \hline
		\cite{nikolaos95}(1994) & Pre-segmented Data + RNN at character level  \\ \hline
		\cite{chen99} (1999) & A dynamic programming based approach to find best path \\ \hline
		\cite{Toselli2004}(2004) & Stochastic finite-sate transducers + HMM \\ \hline
		\cite{graves2006}(2006) & CTC techniques which is widely in HTR introduced in speech recognition task. \\ \hline  
		\cite{dreuw2011}(2011) & Multi layer perceptron + HMM \\ \hline
		\cite{zamora2011}(2011) & Artificial Neural Network + HMM\\ \hline
		\cite{Liwicki2012}(2012) & BLSTM + CTC based recognition and end-to end training. \\ \hline
		\cite{cirricum2013}(2013) & MDLSTM + Covolution Layers + Fully Connected Network (FCN) + CTC. \\ \hline
		\cite{kozielski2013}(2013) &  LSTM-RNN tandem HMM system. \\ \hline
		\cite{doetsch2014}(2014) & LSTM-RNN and accelerating back propagation through time using mini batching. \\ \hline
		\cite{pham2014}(2014) & MDLSTM + CTC + Dropout. it helps to reduce overfitting and part of every model usually.\\ \hline
		\cite{bluche2016}(2016) & MDLSTM-RNN + Vertical Attention to recognize text at line level and paragraph level.\\ \hline
		
		\cite{Voigtlaender2016}(2016)& Large MDLSTM-RNN + CTC.\\ \hline
		\cite{Doetsch2016}(2016) & CNN + BLSTM + CTC. \\ \hline
		\cite{puigcerver2017}(2017) & CNN+ BLSTM + CTC. Stacking of BLSTM layers giving MDLSTM equvilent performance.\\ \hline
		\cite{bluche2017}(2017) & Gated-CNN + BLSTM.\\ \hline 
		\cite{Castro2018}(2018) & MDLSTM + HMM to boost recognition speed and improve accuracy.\\ \hline 
		\cite{Dutta2018}(2018) & CNN-RNN hybrid model + pre-training with synthetic data and transformations.\\ \hline
		\cite{Chowdhury2018}(2018) & CNN + Map2Seq + BLSTM Encoder + Attention + LSTM Decoder.\\ \hline 
		\cite{Michael2019}(2019) & Seq2Seq Encoder and Decoder model with Attention.\\ \hline 
		\cite{Kang2020}(2020) &  Transformers (CNN feature extractor + multi-headed self-attention layers). Non recurrent model.\\\hline 
		\cite{flor2020}(2020) & CNN + Gated-CNN + BLSTM + CTC on text line images of benchmarked dataset.\\ \hline
		\cite{Coquenet2022}(2022) & FCN+LSTM+Attention of paragraph text recognition with internal line segmentation.\\\hline        
	\end{tabular}
	\label{table:tab1}
\end{table}

\section{System Design}
\begin{figure}[t]
	\includegraphics [width=\linewidth, height=8cm]{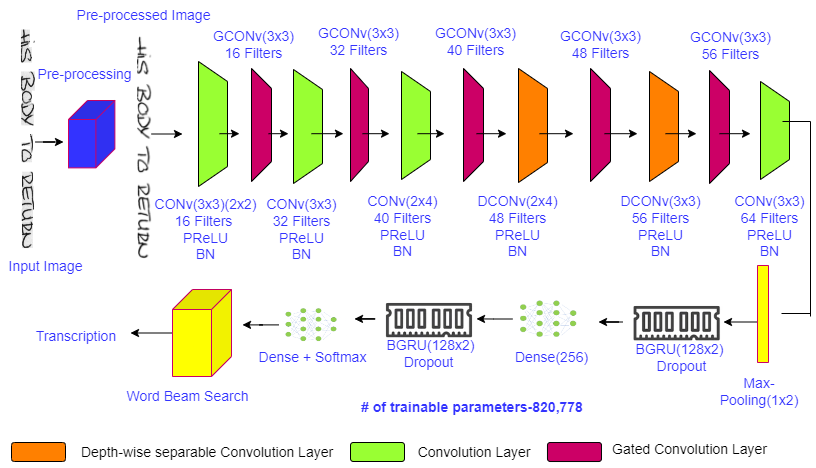}
	\caption {Proposed Architecture}	
	\label{figure:fig-2}
\end{figure}

In this section, we have presented about the basic building blocks of the system proposed. The Figure \ref{figure:fig-2} shows the   
detailed view of our proposed architecture. Here, a text line image $I$ is given as input. The preprocessing  is performed to reduce the noise in the image as well as for improving accuracy. This image is  processed by series of convolutional, gated convolutional and depthwise  separable convolutional layers to extract features from it. The extracted features are propagated by series of by gated recurrent layers. The output of recurrent unit is processed by dense layer and character probability is obtained by applying the softmax as a last layer. The CTC is used to calculate the loss during training of the model. The WBS decoder is used in testing of the model. This model is similar at design level to \cite{flor2020}, except for the middle two depthwise convolutional layer, which helps in reducing total number of trainable parameters. Overview of the components of the system has been discussed in followoing subsections, 

\subsection{Convolutional Layer}
One of the abilities of the networks that has convolutional layer that make it popular among image recognition tasks is the  ablility to detect more abstract features, as the network become deeper and able to extract features regardless of their position in the image \cite{Albawi2017}. Pooling is used alongside of the convolution to downsample the image, thus, have less number of parameters to be taken care at next layer. Further, gates are introduced in convolutional layers to extract a larger context \cite{bluche2017}. So, as deep learning techniques progresses research community focuses on identifying model that helps in reduction in training time of HTR model with less number of parameters.\par
\begin{figure}[h]
 	\includegraphics [width=\linewidth, height=7cm]{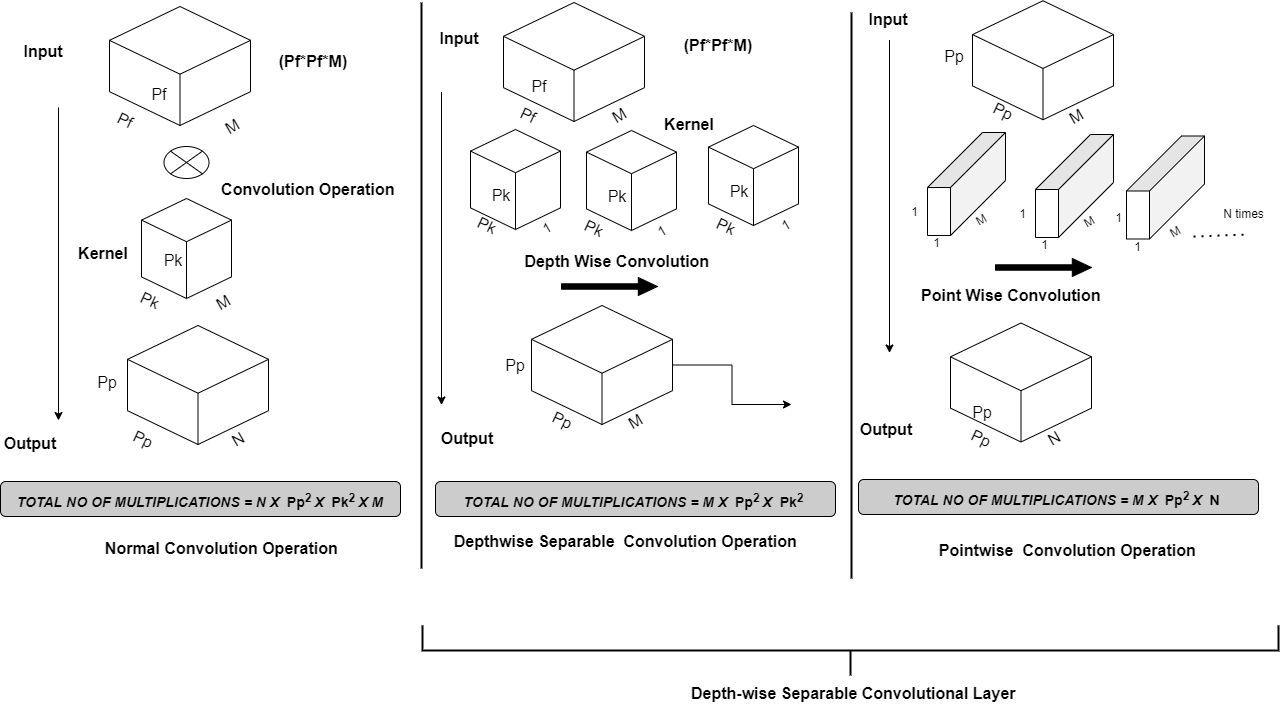}
 	\caption {Comparison between standard convolution and depthwise  separable convolution}	
 	\label{figure:fig-3}
 \end{figure}

 Depthwise  separable convolutional layer is a variation of convolutional layer in which convolution operation is performed to single channel at a time rather than standard convolution that work on all channels. In figure [\ref{figure:fig-3}], Input data is of size, (P\textsubscript{f}$\times$P\textsubscript{f}$\times$M) where P\textsubscript{f}$\times$P\textsubscript{f} is image size and $\textit{M}$ is the number of channels. Assume, we have N filters of size ( P\textsubscript{k}$\times$P\textsubscript{k}$\times$M). Normal convolutional operation produces the output size of (P\textsubscript{p}$\times$P\textsubscript{p}$\times$N) with total number of operations as ($N$$\times$P\textsubscript{p}\textsuperscript{2}$\times$P\textsubscript{p}\textsuperscript{2}$\times$M). While in the case of depthwise  separable convolution with input of (P\textsubscript{f}$\times$P\textsubscript{f}$\times$M) and Kernal size of ( P\textsubscript{k}$\times$P\textsubscript{k}$\times$1) the total number of multiplicative operations are $M \times P p^{2} \times\left(P k^{2}+N\right)$ and where $M \times P k^{2} \times P p^{2}$ operations contributed by depthwise convolution and $M \times P p^{2} \times N$ added by point wise convolution. 
 
\subsection{Recurrent Layer}
Recurrent layers are used to remember part of sequence in a sequence learning task. It is able to utilise the context information of input under processing. In this study, we are using BGRU, since it has lesser parameter than BLSTM, it consist of two GRUs, one for taking input in forward direction and, other for taking input in backward direction \cite{cho2014}.

\section{Experimental Setup and Results}
In this section, the experimental setup of the present work has been presented. The recognition model and WBS decoder have been taken from \cite{flor2020}\footnote[1]{https://github.com/arthurflor23/handwritten-text-recognition} and \cite{Scheidl2018}\footnote[2]{https://github.com/githubharald/CTCWordBeamSearch} respectively. A HTR system is a unique combination of many hyperparameters that needs to be looked upon while designing it and these parameters might be unique to that system only such as the minimum number of image size of input line image(1024x128x1 (Height $\times$ Width $\times$ Channel)), the number of convolution layers, kernel and filter size of each layer, number and position of max-pooling layer with suitable kernel, the total number of recurrent layer, choice of type of recurrent layer (LSTM/BLSTM/MDLSTM/BGRU) and number of unit it should contain, choice of activation function to have non linearity, size of batches, total number of epochs, rate and place of dropout (it is used to regularize the network and reduce overfitting \cite{pham2014}), learning rate, choice of data augmentation while training, stopping criteria of training and choice of optimizer used for training.    
\subsection{Datasets}
Benchmarked datasets such as IAM \cite{IAM}, George Washington (GW) \cite{gw} have been used in this study to evaluate proposed architecture.
\subsubsection{IAM Dataset}
The IAM dataset contains English handwritten forms that is used to train and text HTR models. It is obtained from LOB corpus. It was first published in ICDAR 1999 and currently 3.0 version is for public access including  657 writers and 1539 scanned pages. It has 13353 isolated and  labelled text lines. Table \ref{table:tab2} shows the train, test and validation split used in this study.
\subsubsection{GW Dataset}
This dataset is created from George Washington Papers in English at the Library of Congress. It consist of 20 pages 656 text lines 4894 word instances and 82 unique character. The availability of less data make this dataset challenging for HTR task.Table \ref{table:tab2} shows the train, test and validation split used in this study.

\begin{table}[]
	\centering
	\caption{Train, validation and test splits of benchmarked datsets}
	\begin{tabular}{ccccc}
	\textbf{SNo.} &\textbf{Dataset}                                     & \textbf{Train Images} & \textbf{Validation Images} & \textbf{Test Images} \\ \hline
	1 &	IAM Dataset (\# of characters=79)& 6,161                                    & 900                                    & 1,861                                \\ \hline
	2& GW Dataset (\# of characters=82)  &325                                     & 168       & 163                                  \\ \hline
	\end{tabular}
\label{table:tab2}
\end{table}
\subsection{Preprocessing}
\begin{figure}[t]
	\includegraphics [width=\linewidth, height=4cm]{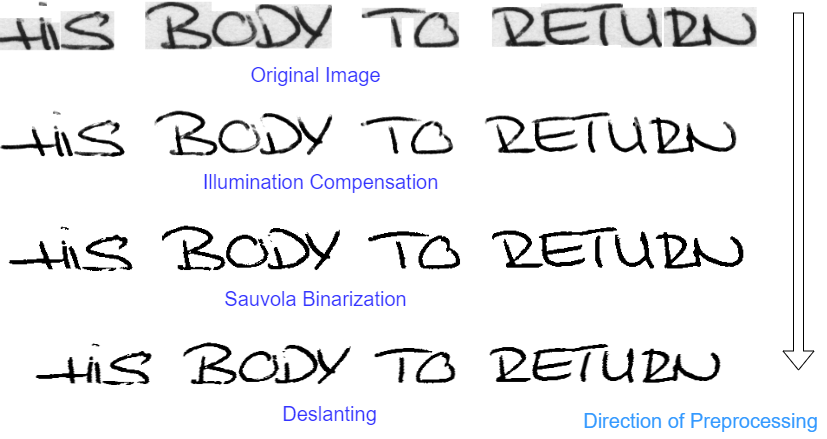}
	\caption {Pre-processing steps}	
	\label{figure:fig-1}
\end{figure}

Preprocessing steps are applied to reduce the noise in the raw data and make it more convenient  for NN models to learn. Handwritten text document specially historical handwritten documents are of poor quality and in degraded stage. Pre-processing techniques such as binarization and normalization has been applied to remove noise in text images. Figure \ref{figure:fig-1} shows the part of text line image of IAM dataset before and after applying preprocessing.
\vspace{-1em}
\subsubsection{Illumination Compensation} With the help of illumination compensation technique the uneven light distribution of the documents have been balanced \cite{CHEN2012}. A light balanced image is produced following the steps such as contrast enhancement, edge detection, text location and light distribution.
\vspace{-1em}
\subsubsection{Binarization} Binarization is applied as a part of preprocessing on the most of the HTR systems. A bi-level image is obtained in this process. It reduces the computational load of the system as compared to 256 levels of grey scale or colour image. In this study, Sauvola binarization is used \cite{SAUVOLA2000}. It uses hybrid approach for deciding threshold value and taking the region properties of document classes into consideration while binarizing.
\vspace{-1em}
\subsubsection{Deslanting} It is a normalization technique. These techniques are used to address the variability among different authors or even same author writing style by making them resemble as close as possible. In this study, a slant and slope removal technique is used to normalize text data by utilising density probability distributions graphs \cite{VINCIARELLI2001}.

\subsection{Evaluation Metric}
We have used evaluation metric that is CER and WER to compare our system with other systems. CER is the number of operations required at character level to transform the ground truth into the output of the recognition model as followed in Eq. \ref{eq:1} where $S_{char}$ is number of substitutions, $D_{char}$ is number of deletions and $I_{char}$ is number of insertions required at character level and $N_{char}$ is total number of characters in ground truth word.
\begin{equation}\label{eq:1}
	CER=\frac{S_{char}+D_{char}+I_{char}}{N_{char}}
\end{equation} 
\subsection{Training Details}
\vspace{-0.59em}
In this section, we have discussed the training process using algorithm \ref{algorithm:algo1} and testing process using algorithm \ref{algorithm:algo2} of the present HTR system in a line by line manner as follows,\\\\\\
	\begin{algorithm}[H]
		\SetAlgoLined
		\KwIn{Text line images $I_1,I_2,... I_n$ and their transcriptions $y_1,y_2,... y_n$}
		\KwResult{Trained model weights on minimizing the validation loss}
		
		$epochs$=1000, $batch$=16, $lr$=0.001, $stop\_tolerence$=20, $reduce\_tolerence$=15 \tcp*{initiliaze the training parameters}
		init model() \tcp*{initialize the model framework}
		\For{$i$=1 \KwTo $batch$}{
			augmentImage($I_i$)\tcp*{Augment text line images}
			$\hat{y_i}$=model($I_i$) \tcp*{Image prediction from model}
			$\delta$\textsubscript{ctc}+=L\textsubscript{ctc}($y_i$,$\hat{y_i}$)\tcp*{compute CTC loss}
		}
	Backward(	$\delta$\textsubscript{ctc}) \tcp{Updated model weights using back propagation}
	\caption{Training Process}
	\label{algorithm:algo1}
	\end{algorithm}
\vspace{-2em}
\subsubsection{Explanation-} We will discuss training process in line by line manner as follows,\\
\begin{tabular}{lp{11cm}}
	
	\textbf{Line 1:}-&  Initialize the model parameters required for training such as, total number of epochs, batch size learning rate, early stopping criteria for training and reduce learning rate on plateau (factor 0.2).\\
	\textbf{Line 2:}-& Preparation of the model architecture.\\
	\textbf{Line 3-4:}-& For each image in the training batch random data  augmentation is performed such as random  morphological and displacement
	transformations that includes rotation, resizing, erosion and dilation.\\
	\textbf{Line 5:}-& Augmented image, obtained in earlier step is fed into NN the model.\\
	\textbf{Line 6-7:}-& For each text line image CTC loss is calculated and combined.\\
	\textbf{Line 8:}-& The model weights are updated based on loss value using back propagation through time algorithm. This training is continued until maximum number of epochs reached or stopping criteria has been met whichever is earlier. 
\end{tabular}

\begin{algorithm}[H]
	\SetAlgoLined
	\KwIn{ Text line image $I$, $D_{test_{corpus}}$, $D_{chars}$, $D_{wordchars}$}
	\KwResult{Transcription of the image along with CER and WER}
	$BW$=50, $mode$='NGrams', $smooth$=0.01 \tcp*{initialize the WBS decoding parameters}
	initModel() \tcp*{Loading of the trained model}
	out=predict($I$) \tcp*{Predict the text of image}
	swapaxis(out) \tcp*{Swapping of the output axis to make it suitable for WBS decoder}
	$\hat{y}$= WBS($BW$,$mode$, $smooth$=0.01,$D_{test_{corpus}}$, $D_{chars}$, $D_{wordchars}$) \tcp*{Applying WBS decoding algorithm}
	$CER$,$WER$= accuracy($y$,$\hat{y}$) \tcp*{compute accuracy}
\caption{Prediction Process}
\label{algorithm:algo2}
\end{algorithm}	
\subsubsection{Explanation-} We will discuss prediction process in line by line manner as follows,\\
\begin{tabular}{lp{11cm}}
	
	\textbf{Line 1:}-&  Input of the model is text line image $I$. First, we initialize the WBS decoding parameters such as beam width to 50, mode of algorithm to 'NGrams' and  smootihing factor as 0.01.\\
	\textbf{Line 2:}-& Preparation of the model.\\
	\textbf{Line 3:}-& The model takes input text line image $I$ and produce  probabilities of each character at each time step.\\
	\textbf{Line 4:}-& The output matrix dimensions are swapped as per predefined input accepted by WBS decoder.\\
	\textbf{Line 5:}-& Computation of the text using WBS decoding algorithm.\\
	\textbf{Line 6:}-& calculation of CER and WER.
\end{tabular}    	
\vspace{-1em}
\subsection{Results and Comparison}
As discussed, we have used benchmarked datasets and compare this with other state-of-the art techniques. For recognition, we have used the implementation of the base model of \cite{flor2020} and for WBS decoding we have used the implementation of \cite{Scheidl2018}. The total number of trainable parameters in flor et al.\cite{flor2020} are 822,770, while, in this study total number of trainable parameters are 820,778 due to use of  depthwise separable convolutional layers (DCNN in table \ref{table:tab3})  instead of standard convolutional layers. Table \ref{table:tab3} and Table \ref{table:tab4} summarizes our findings. We are able to achieve 3.84\% CER and 9.40 \%WER on IAM dataset and 4.88\% CER and 14.56\% WER on GW dataset.

\begin{table}[]
	\caption{Comparision of present work with other state-of-the-art works on IAM Dataset}
	\centering 
	\begin{tabular}{lllll}
		\textbf{SNo.}	& \textbf{Reference}  & \textbf{ Method/Technique }    & \textbf{CER}         & \textbf{WER}         \\ \hline
		1 & Puigcerver et al. \cite{puigcerver2017}	& CNN + LSTM + CTC       & 4.4                    & 12.2                     \\ \hline
		2 & Chowdhury et al.\cite{Chowdhury2018} & CNN + BLSTM + LSTM        & 8.1                   & 16.7                     \\ \hline
		3 & Michael et al. \cite{Michael2019} & CNN + LSTM+ Attention  & 4.87                  & -                  \\ \hline
		4 &  Kang et al. \cite{Kang2020} & Transformer      & 4.67                 & 15.45                \\ \hline
		5 & Yousef et al.\cite{YOUSEF2020} & CNN + CTC       & 4.9     & -               \\ \hline
		6 & Flor et al. \cite{flor2020} & CNN + BGRU + CTC  & 3.72 & 11.18 \\ \hline
		7 &	 Present Work & CNN + DCNN + BGRU+ CTC       & \textbf{3.84}              & \textbf{9.40}              \\ \hline

	\end{tabular}
	\label{table:tab3}
\end{table}
\begin{table}[]
	\caption{Comparision of present work with other state-of-the-art works on GW Dataset}
	\centering 
	\begin{tabular}{lllll}
		\textbf{SNo.} &	\textbf{Reference} & \textbf{Method/Technique}  & \textbf{CER}         & \textbf{WER}         \\ \hline
		1 & Toledo et al. \cite{Toledo2017} & CNN + BLSTM + CTC   &  7.32                   & -                   \\ \hline
		2 & Almazan et al. \cite{Almazan2014}  &   Word Embedding  &  17.40                  &    -          \\ \hline
		3 & Fischer et al. \cite{Fischer2012} &   HMM + RNN   &  20  & -   \\ \hline
		4 & Present Work &  CNN + BGRU + CTC  &  \textbf{4.88}                & \textbf{14.56}                   \\ \hline
		
	\end{tabular}
	\label{table:tab4}
\end{table}

\section{Discussion}
\begin{figure}[!htbp]
	\includegraphics [width=\linewidth, height=10cm]{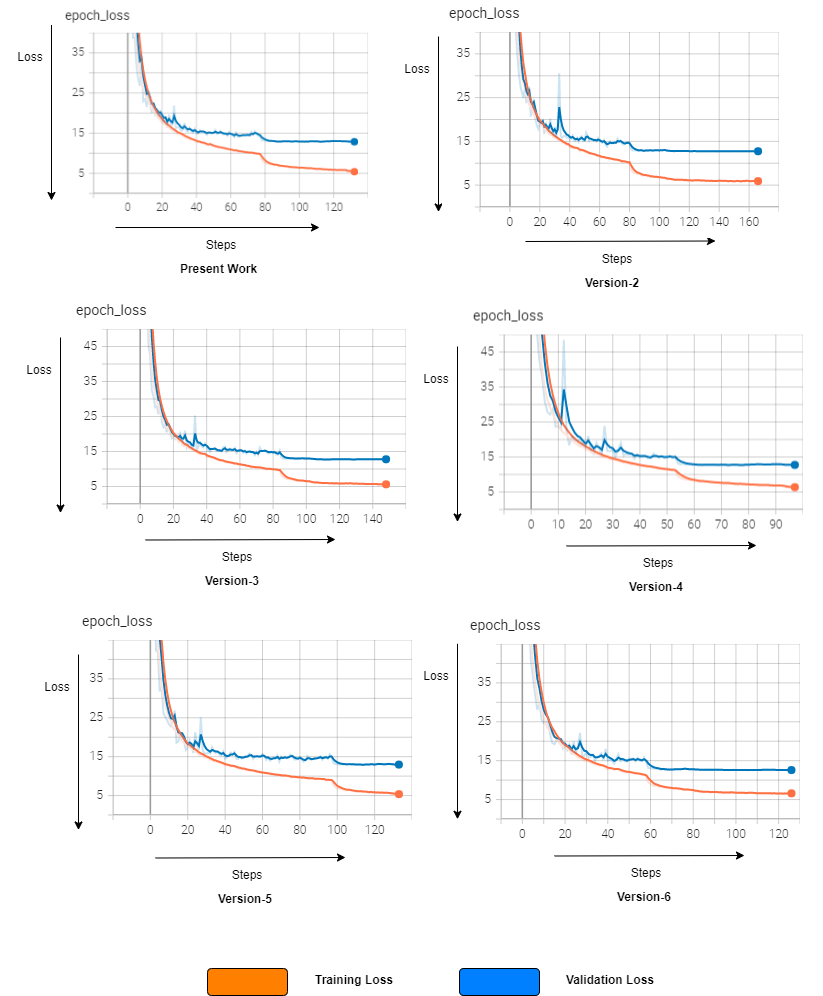}
	\caption {Traning and Validation loss curves of Proposed system and variations studied}	
	\label{figure:fig-4}
\end{figure}  

\begin{table}[!htbp]
	\centering
	\ra{1.2}
	\label{tab:table5}
	\caption{Variations of layers in proposed model}
	\begin{tabular}{lll}
		\multicolumn{1}{c}{Name} & \multicolumn{1}{c}{Model} &\multicolumn{1}{c}{ \# of trainable parameters} \\ \hline
		\multicolumn{1}{c}{Present work} 	& \multicolumn{1}{c}{ C--C--C--D--D--C}     &    	\multicolumn{1}{c}{ 820,778}                  \\ \hline
		\multicolumn{1}{c}{ Version-2 }	&   \multicolumn{1}{c}{ D--D--D--D--D--D}  &           	\multicolumn{1}{c}{ 818,492}               \\ \hline
		\multicolumn{1}{c}{ Version-3 }	& \multicolumn{1}{c}{ C--C--C--D--D--D}      &   	\multicolumn{1}{c}{ 821,122}                       \\ \hline
		\multicolumn{1}{c}{ Version-4 }	&   \multicolumn{1}{c}{ C--D--C--D--C--D}    &    	\multicolumn{1}{c}{ 819,682}                \\ \hline
		\multicolumn{1}{c}{ Version-5}	&   \multicolumn{1}{c}{ C--C--D--D--D--D}     &    	\multicolumn{1}{c}{820,386}                       \\ \hline
		\multicolumn{1}{c}{ Version-6}	&    \multicolumn{1}{c}{ C--D--D--D--D--D}    &      	\multicolumn{1}{c}{ 818,610 }                    \\ \hline
	\end{tabular}
\end{table}
Here, we have discussed various changes in the layers while studying this model. While doing variations, we did not change the settings of the configuration of gated convolutional layer and recurrent layer.
We only varied the positions of the convolutional and depthwise  separable convolution layer, thus we study 6 different variants of proposed architecture. The table C represents standard convolution and D represents depth wise  separable convolution. Figure \ref{figure:fig-4} represents the training and validation loss curve of each of model variants presented in table \ref{tab:table5}. It is evident from the figure \ref{figure:fig-4} that variation in the number of depthwise separable convolutional layers does not effect much in the model performance. \par For statistical analysis, since each iteration of NN training is independent of each other, we have executed thirty training iterations of each variation of table 5 on the IAM dataset and used one-way Analysis Of Variance (ANOVA) test \cite{scheffe1999}  with 5 \% significance. As null hypothesis, we considered $H_0:\mu_1=\mu_2=\mu_3=\mu_4=\mu_5=\mu_6$ and alternative hypothesis $H_1:$ at least one of the $\mu_i$ (where $i=1$ to 6) is different. In this, $\mu_1, \mu_2, \mu_3, \mu_4, \mu_5, \mu_6$ are the different rows of table \ref{tab:table5}. We analyse for both CER and WER. The obtained p-value is 0.68614, which is greater than 0.05; thus, we failed to reject the null hypothesis.
\section{Conclusion}
 We have presented a text line handwritten text recognition model using state-of-the-art approaches in each module of the HTR system. We have used depthwise convolution layers to reduce the number of parameters for training  and obtained results similar to state-of-the-art techniques. Various different type of model architectures are discussed by changing the number and position of depth-wise separable convolution layers. The ANOVA statistical test has been performed on these models to show their performance similarity irrespective of the model architecture. We have also implemented the WBS algorithm at the decoding step while testing, which improves the obtained results.

\subsubsection{Acknowledgements} This research is funded by Government of India, University Grant Commission, under Junior Research Fellowship
scheme.

 \bibliographystyle{splncs03unsrt}
 \bibliography{bibliography}

\begin{thebibliography}{10}
\providecommand{\url}[1]{\texttt{#1}}
\providecommand{\urlprefix}{URL }

\bibitem{Kumari2022}
Kumari, L., Sharma, A.: A review of deep learning techniques in document image
  word spotting. Archives of Computational Methods in Engineering  29(2),
  1085--1106 (Mar 2022)

\bibitem{Scheidl2018}
Scheidl, H., Fiel, S., Sablatnig, R.: Word beam search: A connectionist
  temporal classification decoding algorithm. In: 2018 16th International
  Conference on Frontiers in Handwriting Recognition (ICFHR). pp. 253--258
  (2018)

\bibitem{chen99}
Chen, W.T., Gader, P., Shi, H.: Lexicon-driven handwritten word recognition
  using optimal linear combinations of order statistics. IEEE Transactions on
  Pattern Analysis and Machine Intelligence  21(1),  77--82 (1999)

\bibitem{bellman2015}
Bellman, R.E., Dreyfus, S.E.: Applied dynamic programming, vol. 2050. Princeton
  university press (2015)

\bibitem{2002vin}
Vinciarelli, A.: A survey on off-line cursive word recognition. Pattern
  Recognition  35(7),  1433--1446 (2002)

\bibitem{Toselli2004}
Toselli, A., Juan, A., González, J., Salvador, I., Vidal, E., Casacuberta, F.,
  Keysers, D., Ney, H.: Integrated handwriting recognition and interpretation
  using finite-state models. International Journal of Pattern Recognition and
  Artificial Intelligence  18(4),  519 – 539 (2004)

\bibitem{SANCHEZ2019}
Sánchez, J.A., Romero, V., Toselli, A.H., Villegas, M., Vidal, E.: A set of
  benchmarks for handwritten text recognition on historical documents. Pattern
  Recognition  94,  122--134 (2019)

\bibitem{zamora2011}
Espana-Boquera, S., Castro-Bleda, M., Gorbe-Moya, J., Zamora-Martinez, F.:
  Improving offline handwritten text recognition with hybrid hmm/ann models.
  IEEE Transactions on Pattern Analysis and Machine Intelligence  33(4),
  767--779 (2011)

\bibitem{dreuw2011}
Dreuw, P., Doetsch, P., Plahl, C., Ney, H.: Hierarchical hybrid mlp/hmm or
  rather mlp features for a discriminatively trained gaussian hmm: A comparison
  for offline handwriting recognition. In: 2011 18th IEEE International
  Conference on Image Processing. pp. 3541--3544 (2011)

\bibitem{Toselli2015}
Toselli, A.H., Vidal, E.: Handwritten text recognition results on the bentham
  collection with improved classical n-gram-hmm methods. In: Proceedings of the
  3rd International Workshop on Historical Document Imaging and Processing. p.
  15–22 (2015), \url{https://doi.org/10.1145/2809544.2809551}

\bibitem{doetsch2014}
Doetsch, P., Kozielski, M., Ney, H.: Fast and robust training of recurrent
  neural networks for offline handwriting recognition. In: 2014 14th
  international conference on frontiers in handwriting recognition. pp.
  279--284. IEEE (2014)

\bibitem{kozielski2013}
Kozielski, M., Doetsch, P., Ney, H., et~al.: Improvements in rwth's system for
  off-line handwriting recognition. In: 2013 12th International Conference on
  Document Analysis and Recognition. pp. 935--939. IEEE (2013)

\bibitem{Liwicki2012}
Liwicki, M., Graves, A., Bunke, H.: Neural Networks for Handwriting
  Recognition. Springer Berlin Heidelberg, Berlin, Heidelberg (2012)

\bibitem{nikolaos95}
Bourbakis, N.G., Koutsougeras, C., Jameel, A.: {Handwriting recognition using a
  reduced character method and neural nets}. In: Nonlinear Image Processing VI.
  vol. 2424, pp. 592 -- 601. SPIE (1995)

\bibitem{graves2006}
Graves, A., Fernández, S., Gomez, F., Schmidhuber, J.: Connectionist temporal
  classification: Labelling unsegmented sequence data with recurrent neural
  'networks. vol. 2006, pp. 369--376 (01 2006)

\bibitem{cirricum2013}
Louradour, J., Kermorvant, C.: Curriculum learning for handwritten text line
  recognition

\bibitem{pham2014}
Pham, V., Bluche, T., Kermorvant, C., Louradour, J.: Dropout improves recurrent
  neural networks for handwriting recognition. In: 2014 14th International
  Conference on Frontiers in Handwriting Recognition. pp. 285--290 (2014)

\bibitem{bluche2016}
Bluche, T., Louradour, J., Messina, R.: Scan, attend and read: End-to-end
  handwritten paragraph recognition with mdlstm attention. In: 2017 14th IAPR
  International Conference on Document Analysis and Recognition (ICDAR).
  vol.~01, pp. 1050--1055 (2017)

\bibitem{Voigtlaender2016}
Voigtlaender, P., Doetsch, P., Ney, H.: Handwriting recognition with large
  multidimensional long short-term memory recurrent neural networks. In: 2016
  15th International Conference on Frontiers in Handwriting Recognition
  (ICFHR). pp. 228--233 (2016)

\bibitem{puigcerver2017}
Puigcerver, J.: Are multidimensional recurrent layers really necessary for
  handwritten text recognition? In: 2017 14th IAPR International Conference on
  Document Analysis and Recognition (ICDAR). vol.~01, pp. 67--72 (2017)

\bibitem{shi2017}
Shi, B., Bai, X., Yao, C.: An end-to-end trainable neural network for
  image-based sequence recognition and its application to scene text
  recognition. IEEE Transactions on Pattern Analysis and Machine Intelligence
  39(11),  2298--2304 (2017)

\bibitem{SimpleHtr2018}
Scheidl, H.: Handwritten text recognition in historical document.
  diplom-Ingenieur in Visual Computing, Master’s thesis, Technische
  Universität Wien, Vienna (2018)

\bibitem{bluche2017}
Bluche, T., Messina, R.: Gated convolutional recurrent neural networks for
  multilingual handwriting recognition. In: 2017 14th IAPR International
  Conference on Document Analysis and Recognition (ICDAR). vol.~01, pp.
  646--651 (2017)

\bibitem{flor2020}
de~Sousa~Neto, A.F., Bezerra, B.L.D., Toselli, A.H., Lima, E.B.: Htr-flor: A
  deep learning system for offline handwritten text recognition. In: 2020 33rd
  SIBGRAPI Conference on Graphics, Patterns and Images (SIBGRAPI). pp. 54--61
  (2020)

\bibitem{Doetsch2016}
Doetsch, P., Zeyer, A., Ney, H.: Bidirectional decoder networks for
  attention-based end-to-end offline handwriting recognition. In: 2016 15th
  International Conference on Frontiers in Handwriting Recognition (ICFHR). pp.
  361--366 (2016)

\bibitem{Coquenet2022}
Coquenet, D., Chatelain, C., Paquet, T.: End-to-end handwritten paragraph text
  recognition using a vertical attention network. IEEE Transactions on Pattern
  Analysis and Machine Intelligence pp. 1--1 (2022)

\bibitem{Castro2018}
Castro, D., L.~D.~Bezerra, B., Valença, M.: Boosting the deep multidimensional
  long-short-term memory network for handwritten recognition systems. In: 2018
  16th International Conference on Frontiers in Handwriting Recognition
  (ICFHR). pp. 127--132 (2018)

\bibitem{Dutta2018}
Dutta, K., Krishnan, P., Mathew, M., Jawahar, C.: Improving cnn-rnn hybrid
  networks for handwriting recognition. In: 2018 16th International Conference
  on Frontiers in Handwriting Recognition (ICFHR). pp. 80--85 (2018)

\bibitem{Chowdhury2018}
Chowdhury, A., Vig, L.: An efficient end-to-end neural model for handwritten
  text recognition (2018), \url{https://arxiv.org/abs/1807.07965}

\bibitem{Michael2019}
Michael, J., Labahn, R., Gruning, T., Zollner, J.: Evaluating
  sequence-to-sequence models for handwritten text recognition. pp. 1286--1293
  (09 2019)

\bibitem{Kang2020}
Kang, L., Riba, P., Rusiñol, M., Fornés, A., Villegas, M.: Pay attention to
  what you read: Non-recurrent handwritten text-line recognition (2020),
  \url{https://arxiv.org/abs/2005.13044}

\bibitem{Albawi2017}
Albawi, S., Mohammed, T.A., Al-Zawi, S.: Understanding of a convolutional
  neural network. In: 2017 International Conference on Engineering and
  Technology (ICET). pp. 1--6 (2017)

\bibitem{cho2014}
Cho, K., van Merrienboer, B., Bahdanau, D., Bengio, Y.: On the properties of
  neural machine translation: Encoder-decoder approaches (2014),
  \url{https://arxiv.org/abs/1409.1259}

\bibitem{IAM}
Marti, U.V., Bunke, H.: A full english sentence database for off-line
  handwriting recognition. In: Proceedings of the Fifth International
  Conference on Document Analysis and Recognition. p. 705. ICDAR '99, IEEE
  Computer Society, USA (1999)

\bibitem{gw}
Fischer, A., Keller, A., Frinken, V., Bunke, H.: Lexicon-free handwritten word
  spotting using character hmms. Pattern Recognition Letters  33(7),  934--942
  (2012),
  \url{https://www.sciencedirect.com/science/article/pii/S0167865511002820},
  special Issue on Awards from ICPR 2010

\bibitem{CHEN2012}
Chen, K.N., Chen, C.H., Chang, C.C.: Efficient illumination compensation
  techniques for text images. Digital Signal Processing  22(5),  726--733
  (2012),
  \url{https://www.sciencedirect.com/science/article/pii/S1051200412000826}

\bibitem{SAUVOLA2000}
Sauvola, J., Pietikäinen, M.: Adaptive document image binarization. Pattern
  Recognition  33(2),  225--236 (2000),
  \url{https://www.sciencedirect.com/science/article/pii/S0031320399000552}

\bibitem{VINCIARELLI2001}
Vinciarelli, A., Luettin, J.: A new normalization technique for cursive
  handwritten words. Pattern Recognition Letters  22(9),  1043--1050 (2001),
  \url{https://www.sciencedirect.com/science/article/pii/S0167865501000423}

\bibitem{YOUSEF2020}
Yousef, M., Hussain, K.F., Mohammed, U.S.: Accurate, data-efficient,
  unconstrained text recognition with convolutional neural networks. Pattern
  Recognition  108,  107482 (2020),
  \url{https://www.sciencedirect.com/science/article/pii/S0031320320302855}

\bibitem{Toledo2017}
Toledo, J.I., Dey, S., Fornes, A., Llados, J.: Handwriting recognition by
  attribute embedding and recurrent neural networks. In: 2017 14th IAPR
  International Conference on Document Analysis and Recognition (ICDAR).
  vol.~01, pp. 1038--1043 (2017)

\bibitem{Almazan2014}
Almazan, J., Gordo, A., Fornes, A., Valveny, E.: Word spotting and recognition
  with embedded attributes. IEEE Transactions on Pattern Analysis and Machine
  Intelligence  36(12),  2552--2566 (2014)

\bibitem{Fischer2012}
Fischer, A.: Handwriting recognition in historical documents. Ph.D. thesis,
  Verlag nicht ermittelbar (01 2012)

\bibitem{scheffe1999}
Scheffe, H.: The analysis of variance, vol.~72. John Wiley \& Sons (1999)

\end{thebibliography}

\end{document}